\documentclass[final,5p,times,twocolumn]{elsarticle}

\def\tsc#1{\csdef{#1}{\textsc{\lowercase{#1}}\xspace}}
\tsc{WGM}
\tsc{QE}

\usepackage{amssymb}
\usepackage{amsmath}
\usepackage[colorlinks=true,]{hyperref}
\usepackage{tabularx}
\usepackage{booktabs}
\usepackage{multirow}
\usepackage[numbers]{natbib}

\begin{document}
\let\WriteBookmarks\relax
\def\floatpagepagefraction{1}
\def\textpagefraction{.001}

\begin{frontmatter}

\title{Boundary-Aware Test-Time Adaptation for Zero-Shot Medical Image Segmentation}

\author[label1]{Chenlin Xu}
\author[label1]{Lei Zhang}
\author[label1]{Lituan Wang \corref{cor1}}
\author[label1]{Xinyu Pu}
\author[label1]{Pengfei Ma}
\affiliation[label1]{organization={School of Computer Science, Sichuan University},
            city={Chengdu},
            postcode={610065},
            country={China}}
\author[label2]{Guangwu Qian}
\affiliation[label2]{organization={Pittsburgh Institute, Sichuan University},
            city={Chengdu},
            postcode={610065},
            country={China}}
\author[label3] {Zizhou Wang}
\author[label3] {Yan Wang}
\affiliation[label3]{
    organization={Institute of High Performance Computing, Agency for Science, Technology and Research (A*STAR)},
    city={Singapore},
    postcode={138632},
    country={Singapore}}

\cortext[cor1]{
Corresponding author. 
E-mail address: lituanwang@scu.edu.cn(L. Wang).
}

\begin{abstract}
    \textit{Background and Objective:}
    Due to the scarcity of annotated data and the substantial computational costs of model, conventional tuning methods in medical image segmentation face critical challenges. 
    Current approaches to adapting pretrained models, including full-parameter and parameter-efficient fine-tuning, still rely heavily on task-specific training on downstream tasks. 
    Therefore, zero-shot segmentation has gained increasing attention, especially with foundation models such as SAM demonstrating promising generalization capabilities. 
    However, SAM  still faces notable limitations on medical datasets due to domain shifts, making efficient zero-shot enhancement an urgent research goal. \\
    \textit{Methods:}
    To address these challenges, we propose BA-TTA-SAM, a task-agnostic test-time adaptation framework that significantly enhances the zero-shot segmentation performance of SAM via test-time adaptation. This framework integrates two key mechanisms: (1) The encoder-level Gaussian prompt injection embeds Gaussian-based prompts directly into the early feature processing stages of the image encoder, providing explicit guidance for initial representation learning. (2) The cross-layer boundary-aware attention alignment exploits the hierarchical feature interactions within the ViT backbone, aligning deep semantic responses with shallow boundary cues. \\
    \textit{Results:}
    Experiments on four datasets, including ISIC, Kvasir, BUSI, and REFUGE, show an average improvement of 12.4\% in the DICE score compared with SAM's zero-shot segmentation performance. The results demonstrate that our method consistently outperforms state-of-the-art models in medical image segmentation.\\
    \textit{Conclusions:}
    Our framework significantly enhances the generalization ability of SAM, without requiring any source-domain training data. Extensive experiments on publicly available medical datasets strongly demonstrate the superiority of our framework. Our code is available at \underline{https://github.com/Emilychenlin/BA-TTA-SAM}.
\end{abstract}


\end{frontmatter}



\section{Introduction}
\label{sec:introduction}

The impressive zero-shot generalization capabilities of large language models have driven substantial interest in developing foundation models for computer vision tasks \cite{ma2024segment}. The Segment Anything Model (SAM) \cite{kirillov2023segment}, a prompt-based image segmentation model trained on billions of annotated masks, has demonstrated exceptional performance in lots of image domains. By innovatively incorporating prompt mechanisms such as points, bounding boxes, and text, SAM significantly expands its potential applications to adapt to tasks not seen during training \cite{Zheng2024SegmentAC}. 
Unlike traditional models that rely on fixed input-output pairs, SAM incorporates interactive prompt mechanisms, allowing it to generalize to previously unseen objects and scenarios with minimal supervision. 
This flexible prompting paradigm significantly broadens the applicability of SAM to various downstream tasks \cite{Han2023BoostingSA, Zhou2024SAMSPSM, Marinis2024LabelAM}, including interactive annotation, open-vocabulary segmentation, and few-shot learning, thereby making it a promising tool for real-world vision systems.

Although trained on a vast dataset, SAM still faces challenges in medical image segmentation tasks \cite{Chen2023SAMFT}. 
This is primarily attributed to the significant domain shift between its training and target distributions. The training distribution mainly consists of natural images from the SA-1B dataset \cite{kirillov2023segment}, whereas the target medical image domain \cite{Zhang2024SegmentAM} is characterized by complex anatomical structures and subtle texture variations. 
In contrast to natural scenes, Medical images typically require fine-grained boundary delineation and domain-specific context understanding \cite{Xiao2024CATSAMCT}, which are not adequately captured in SAM's original training objectives. 

\begin{figure}[htb]
\centering
\includegraphics[width=1.0\columnwidth]{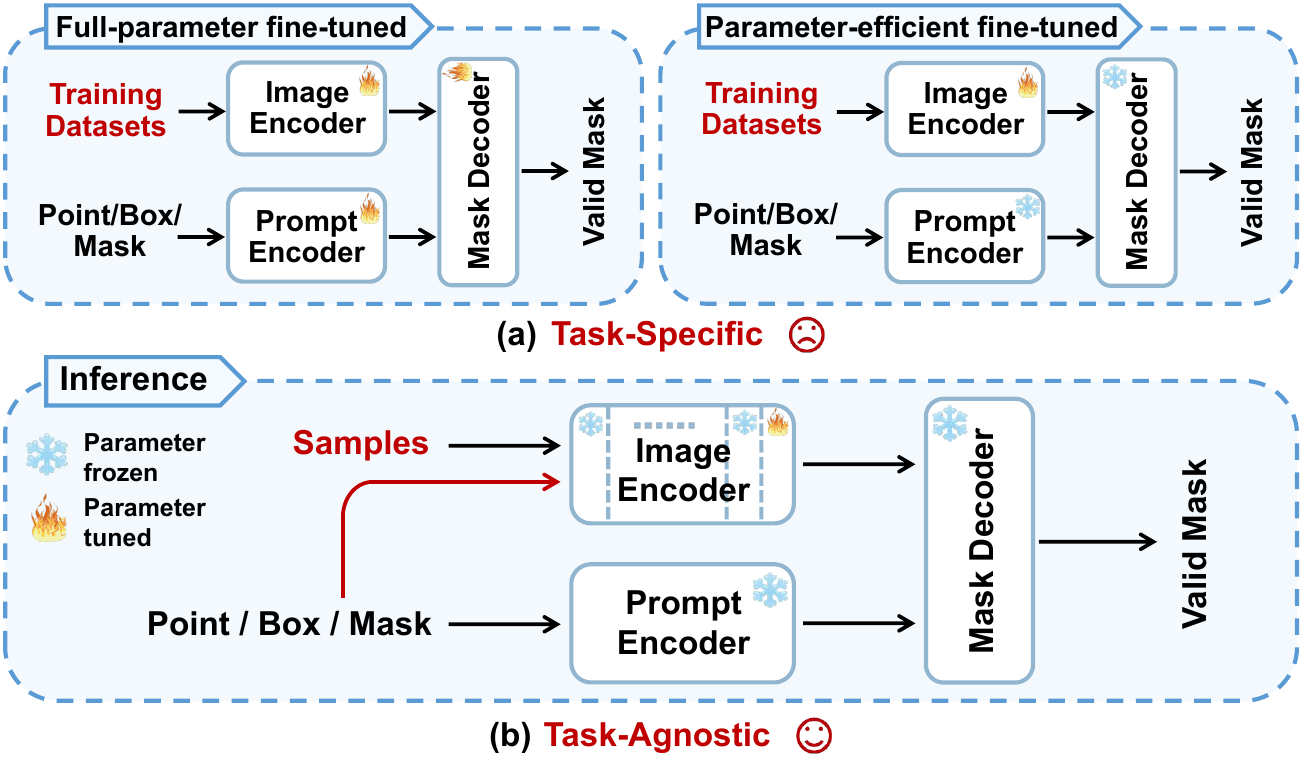} 
\caption{
(a) Existing SAM downstream methods rely on task-specific fine-tuning (full-parameter and parameter-efficient). 
(b) Our BA-TTA-SAM achieves task-agnostic test-time adaptation during inference via prompt injection and boundary-aware alignment, without additional retraining on training datasets. 
}
\label{fig:Frame1}
\end{figure}

Existing paradigms for improving the generalization ability of SAM focus mainly on full-parameter fine-tuning \cite{ma2024segment} and parameter-efficient fine-tuning \cite{zhang2023customized}, as illustrated in Fig.~\ref{fig:Frame1}(a). 
%
The former involve end-to-end fine-tuning of the entire SAM architecture using large-scale, densely annotated medical image segmentation datasets, such as MedSAM \cite{ma2024segment} and SAM-Med2D \cite{Cheng2023SAMMed2D}. Although these approaches can significantly boost performance, they are highly resource intensive, requiring substantial GPU memory, prolonged training times, and domain-specific supervision \cite{Lyu2024MCPMedSAMAP}. This makes them impractical for rapid deployment or low-resource clinical environments \cite{THAKUR2025110615}. 
To address these limitations, parameter-efficient fine-tuning approaches have been proposed, such as WeSAM \cite{10657214} and MSA \cite{WU2025103547}, which aim to adapt SAM by updating only a subset of parameters while freezing the majority of the model. These methods substantially reduce computational overhead and allow for quicker adaptation. 
However, they remain essentially task-specific adaptation methods, relying on fine-tuning for each new dataset or clinical scenario. 
In contrast, test-time adaptation offers a task-agnostic alternative. 
Specifically, by introducing prompts into the image encoder and fine-tuning a small subset of parameters in the last block of the image encoder, our BA-TTA-SAM achieves adaptive inference capability, as shown in Fig.~\ref{fig:Frame1}(b). 

We specify two key limitations in SAM's original architecture that constrain its zero-shot segmentation capability. 
First, prompts interact only with the decoder stage, leaving the encoder to extract image features without any task-specific guidance. Consequently, it often fails to capture lesion-specific cues essential for accurate segmentation, particularly in complex or low-contrast images. 
Moreover, in deeply stacked ViT-based encoders, shallow layers are sensitive to low-level visual cues such as color, texture, and boundaries due to their limited receptive fields \cite{Tao2022PiecewiseLN}, whereas deeper layers encode more abstract semantic features but suffer from excessive attention dispersion caused by large receptive fields. This attention dilution makes precise localization difficult.  

To address these limitations, we propose a novel test-time adaptation framework to enhance SAM's generalization ability. 
Specifically, we introduce two complementary mechanisms that implement the TTA strategy by fully leveraging prompt information and integrating multi-layer features of the ViT-based encoders. 
First, to tackle the encoder prompt deficiency, we introduce an encoder-level Gaussian prompt injection. Unlike direct addition of prompt features, which may overly concentrate attention on discrete points and be sensitive to noise, the Gaussian-based injection produces a smooth, spatially distributed guidance that effectively directs early-stage feature extraction and enhances the encoder's focus on task-relevant regions under domain shifts. 
Second, to mitigate attention dilution, we design a cross-layer boundary-aware attention alignment that leverages the inherent characteristics of ViT layers. By exploiting the complementary focus of shallow and deep layers, this mechanism constrains deep semantic features, preventing over-dispersion of attention and improving boundary localization accuracy.

Extensive experiments on diverse medical image segmentation benchmarks demonstrate that our method can effectively adapt test-time samples and achieve significant performance gains. Specifically, leveraging only the current test samples for optimization, our approach achieves notable improvements over the zero-shot performance of SAM, with an average increase in DICE of 12.4\% in the four datasets.  

The main contribution of this work can be summarized as follows.
\begin{itemize}
    \item We propose a novel test-time adaptation framework for SAM, guided by prompt injection and boundary-aware attention alignment. Our framework addresses the semantic mismatch between natural and medical images by dynamically adapting SAM during inference.
    \item We firstly introduce an encoder-level Gaussian prompt injection strategy into SAM's image encoder. By strengthening its semantic perception, this method enhances SAM's zero-shot segmentation performance in the medical domain during test-time inference.
    \item To encourage better boundary localization and improve segmentation accuracy on diverse medical datasets, we introduce a cross-layer boundary-aware attention mechanism with consistent constraints, aligning boundary features between deep and shallow encoder layers.
\end{itemize}

\section{Related work}
\label{sec:related work}
\subsection{Medical Image Segmentation}
The development of medical image segmentation has been profoundly shaped by foundational architectures in computer vision \cite{XIA2025128740}. Fully Convolutional Networks (FCN) \cite{Shelhamer2014FullyCN} pioneered end-to-end pixel-wise prediction, with its medical instantiation U-Net \cite{Ronneberger2015UNetCN} establishing the encoder-decoder paradigm with skip connections. Subsequent advancements embraced Vision Transformers (ViTs) \cite{Dosovitskiy2020AnII}, where architectures like SETR \cite{Zheng2020RethinkingSS} demonstrated the power of global contextual modeling through pure-transformer feature extraction, while MaskFormer \cite{Cheng2021PerPixelCI} unified segmentation tasks via query-based instance decoding. 
Despite achieving impressive performance on curated medical benchmarks, these models confront significant domain adaptation challenges when deployed on unseen imaging modalities or partially annotated clinical datasets.

Recently, the Segment Anything Model (SAM) \cite{kirillov2023segment} demonstrates strong generalization in natural image segmentation. AS a prompt-driven general segmentation mechanism, it supports multimodal interactions (e.g., points, boxes, text), and trains a highly generalizable vision model on the large-scale SA-1B dataset. 
However, SAM has been found to struggle with medical image segmentation tasks due to the distribution shift between its training data and medical domains, such as low-contrast regions, small lesions, and variations across imaging modalities. 

Current adaptation strategies for SAM can be mainly categorized into full parameter fine-tuning and parameter-efficient lightweight adaptation. 
The former approach, exemplified by MedSAM \cite{ma2024segment}, requires complete fine-tuning on curated medical datasets. While demonstrating improved segmentation accuracy on specific anatomies, this paradigm incurs prohibitive computational costs and risks overfitting to narrow medical domains. 
Therefore, work shifts toward parameter-efficient adaptation, e.g., WeSAM's low-rank weight updates \cite{huang2024segsam}, CON-LoRA-SAM's contrast-enhanced decomposition \cite{Zhong2024ConvolutionML}, and the Medical SAM Adapter bottleneck injection \cite{WU2025103547}. 
Although these methods achieve computational efficiency by selectively updating SAM's parameters through rank-constrained optimization or auxiliary modules, they fundamentally require prior exposure to target medical data distributions during training phases. It requires data set-specific adjustment for individual downstream tasks, limiting their generalizability between various downstream tasks. 
To overcome these limitations, we develop a fully test-time training-free adaptation framework that enhances SAM's generalization and segmentation accuracy under domain shifts.

\subsection{Test-Time Adaptation Paradigm}
Test-time adaptation (TTA) has emerged as a promising solution for enhancing model generalization to unseen domains without requiring access to source data \cite{Valanarasu2022OntheFlyTA, Chen2025TesttimeAF, 10657075}. It particularly well-suited for privacy-sensitive or small-sample scenarios such as medical image analysis. Unlike traditional domain adaptation approaches that rely on extensive fine-tuning and source-target joint training, TTA dynamically adjusts the model during inference, offering high flexibility and efficiency \cite{Chen2023EachTI}. The core idea is to exploit information from the test data itself to adapt model components or internal representations \cite{Wen2023FromDT, Yuan2023RobustTA}, thereby improving performance on domain-shifted inputs without incurring high computational cost.

Recent efforts have explored test-time strategies across various vision tasks. TENT \cite{Wang2020FullyTA} adapts models at test time by minimizing prediction entropy through updating only selected parameters of the normalization layers. DELTA \cite{zhao2023delta} provides a plug-in solution for fully test-time adaptation by correcting unreliable normalization statistics and mitigating class bias, enabling robust performance across diverse test scenarios. T3A \cite{iwasawa2021testtime} addresses distribution shift by constructing a test-time anchor memory from confident predictions to recalibrate classifier prototypes. CoTTA \cite{Wang2022ContinualTD} further improves robustness by employing a stochastic weight averaging mechanism with test-time dropout augmentation. SAR \cite{10204752} introduces sharpness-aware adaptation that optimizes model flatness during inference to enhance generalization. UPL-TTA \cite{10.1007/978-3-031-34048-2_19} adopts uncertainty-guided pseudo-label learning to progressively refine predictions during test time, improving generalization on unlabeled target domains. BayTTA \cite{Sherkatghanad2024BayTTAUM} introduces a Bayesian framework to estimate uncertainty and selectively update model parameters, thereby enhancing robustness under distribution shift. More recently, Wu et al. \cite{10655138} pioneered one-prompt adaptation by introducing learnable hyper-prompts to unify segmentation across different datasets. 

Considering that SAM is a large-scale pre-trained foundation model, it can provide reasonable initial predictions even on unseen data compared with conventional segmentation models. Therefore, we adopt it as our backbone to facilitate faster convergence of TTA approach. 
Unlike previous approaches, our work focuses on optimizing SAM's test-time performance by systematically analyzing the layer-wise characteristics of its image encoder. 

Existing SAM architectures limit prompt influence to the decoder hindering effective early representation and suffer from inconsistent boundary responses across encoder layers.
Our method tackles these issues by proposing the encoder-level gaussian prompt injection and the cross-layer boundary-aware attention alignment, resulting in more robust and accurate segmentation.

\begin{figure*}[htb]
\centering
\includegraphics[width=1.0\textwidth]{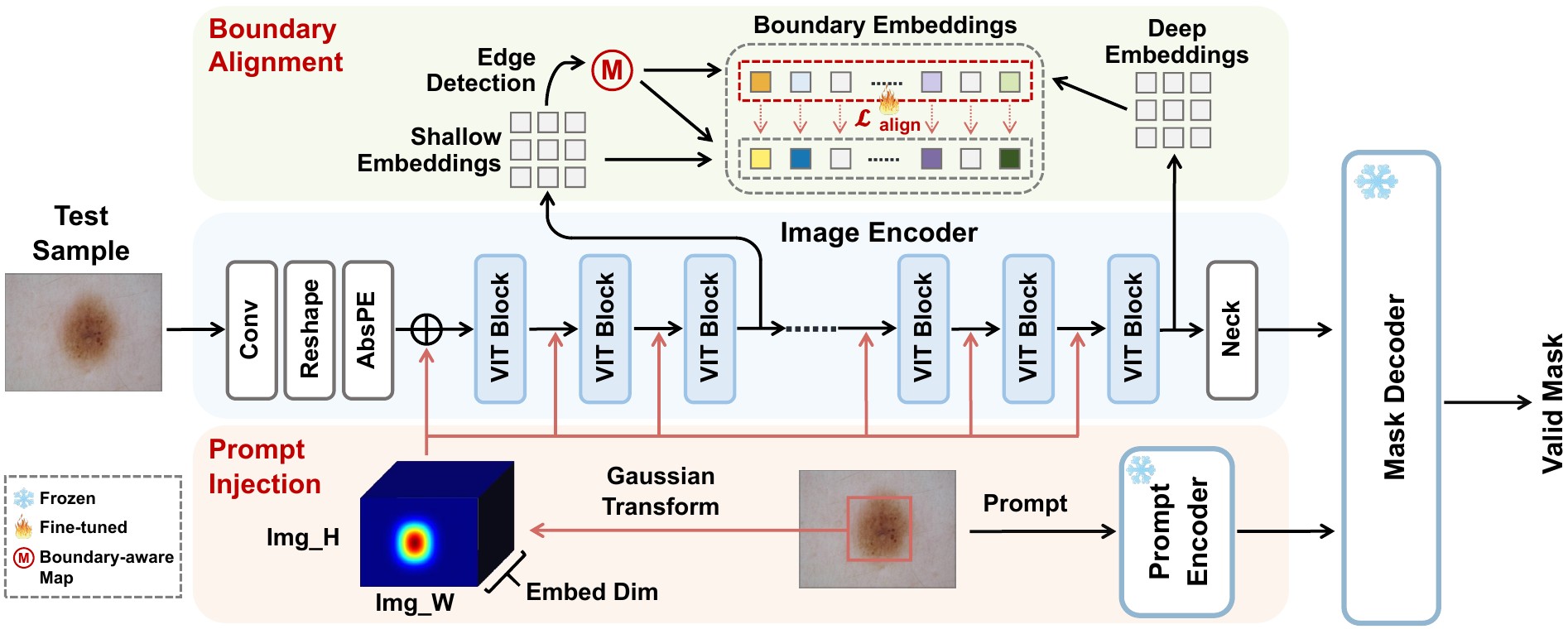} 
\caption{
Overview of our proposed BA-TTA-SAM, consisting of prompt injection and boundary alignment modules. 
The prompt is transformed into Gaussian embeddings and injected into the image encoder at each stage to guide spatial representation learning.
Then, a boundary-aware map is derived from shallow embeddings to enforce shallow–deep consistency via boundary alignment. 
}
\label{fig:Frame2}
\end{figure*}

\section{Methodology}
\label{sec:methodology}
The overall architecture of our proposed BA-TTA-SAM is illustrated in Fig. \ref{fig:Frame2}, which primarily consists of prompt injection and boundary alignment modules.
In this section, we first describe the overall framework of segment-anything model in Section \ref{subsec:3.1overview}. Second, our encoder-level gaussian prompt injection strategy is introduced in Section \ref{subsec:3.2promptinj}. At last, the cross-layer boundary-aware attention alignment is described in Section \ref{subsec:3.3boundaryAlign}.

\subsection{Overview of Segment Anything Model}
\label{subsec:3.1overview}
The Segment Anything Model (SAM) comprises three core components: an image encoder $E_{\text{img}}(I)$, a prompt encoder $E_{\text{prm}}(P)$, and a lightweight mask decoder $D(\mathbf{F}, \mathbf{Q})$. The image encoder $E_{\text{img}}$ is pre-trained using a masked autoencoding objective \cite{He2021MaskedAA} to learn generic visual representations from large-scale unlabeled images. The prompt encoder $E_{\text{prm}}$ transforms various forms of user input (e.g., points, boxes) into embeddings $\mathbf{Q}$ that guide segmentation. The decoder $D$ takes the image features $\mathbf{F}=\{\mathbf{F}^{(l)}\}_{l=1}^{L}=E_{\text{img}}(I)$ and prompt features $\mathbf{Q} = E_{\text{prm}}(P)$ to produce segmentation masks. The full SAM architecture is fine-tuned on the SA-1B dataset \cite{Kirillov2023SegmentA}, which includes over one billion mask annotations, using a hybrid loss function combining focal and Dice loss. During inference, given a test image $I$ and user-provided prompts $P$, the model generates multi-scale mask predictions and a confidence score that estimates segmentation quality. 

In this work, we aim to enhance SAM's adaptability to domain-shifted medical image segmentation through test-time adaptation, which operates directly during inference and requires no access to training data. 
Medical images often exhibit weak contrast and blurred boundaries, leading to insufficient self-attention activation in Vision Transformers and poor lesion–background discrimination. Consequently, SAM's encoder fails to preserve localized attention on lesions, particularly in deeper layers, due to the encoder prompt deficiency. Moreover, while shallow layers in hierarchical ViTs capture edge details, deeper layers lose spatial precision under noisy or ambiguous boundaries because of their large receptive fields, resulting in attention dilution. These two limitations form the core challenges in adapting SAM to domain-shifted medical images, which motivate our proposed framework illustrated in Fig.~\ref{fig:Frame2}.

\subsection{Encoder-Level Gaussian Prompt Injection}
\label{subsec:3.2promptinj}
To address the encoder prompt deficiency of SAM under domain adaptation, an intuitive solution is to inject prompts directly into all encoder layers. However, such a rudimentary strategy may overly concentrate attention on discrete points and remain highly sensitive to noise, which limits its robustness and effectiveness in complex medical scenarios. Considering this limitation, we introduce a Gaussian-based design that provides smooth and spatially distributed guidance. 
Specifically, Gaussian heatmaps provide smooth and spatially distributed guidance that not only mitigates the sensitivity to noise but also balances precise localization with contextual awareness, thereby enhancing the robustness and effectiveness of encoder-level prompt injection. 
Leveraging these properties, we propose Gaussian prompt injection, where prompts are transformed into Gaussian heatmaps and injected across all encoder blocks to continuously guide feature extraction under domain shifts. 

By transforming the prompt into a Gaussian distribution, we obtain a heatmap $\mathcal{H}$ that is injected into the encoder feature hierarchy. Specifically, let $\mathbf{F}^{(l)}$ denote the output feature map of the $l$-th Transformer block within the image encoder $E_{\text{img}}$. The injection is formulated as:

\begin{equation}
    \mathbf{F}^{(l)} = \mathbf{F}^{(l)} + \text{Broadcast}(\mathcal{H})
\end{equation}
where $l$ indexes the Transformer block. Here, $\text{Broadcast}(\mathcal{H})$ denotes expanding the normalized heatmap $\mathcal{H}$ across the feature dimension to match $\mathbf{F}^{(l)}$. 
This encoder-level Gaussian injection modulates intermediate representations in $E_{\text{img}}$, guiding the model to attend to task-relevant regions even under domain shifts.

For a heatmap of size $ H \times W $, we construct a 2D spatial coordinate grid by assigning each pixel its corresponding horizontal and vertical indices. Specifically, for each pixel location ($ i $,$ j $), we define Eq.~\ref{equ:pixel-location}. Each input is mapped to its center coordinate \((x_p, y_p)\). For each such center, we construct a 2D Gaussian distribution $ H_p(i,j) $ to softly highlight the surrounding region. With $N$ such centers, the aggregated heatmap is computed. The corresponding formulations are given as follows:

\begin{equation}
\label{equ:pixel-location}
    \mathbf{X}_{i,j} = j,\quad \mathbf{Y}_{i,j} = i
\end{equation}

\begin{equation}
\label{2d-gaussian}
    H_p(i,j) = \exp\left(-\frac{(\mathbf{X}_{i,j}-x_p)^2 + (\mathbf{Y}_{i,j}-y_p)^2}{2\sigma^2}\right)
\end{equation}

\begin{equation}
\label{equ:raw-heatmap}
    \mathcal{H}_(i,j) = \sum_{p=1}^{N} H_p(i,j)
\end{equation}

\begin{figure}[htb]
\centering
\includegraphics[width=1.0\columnwidth]{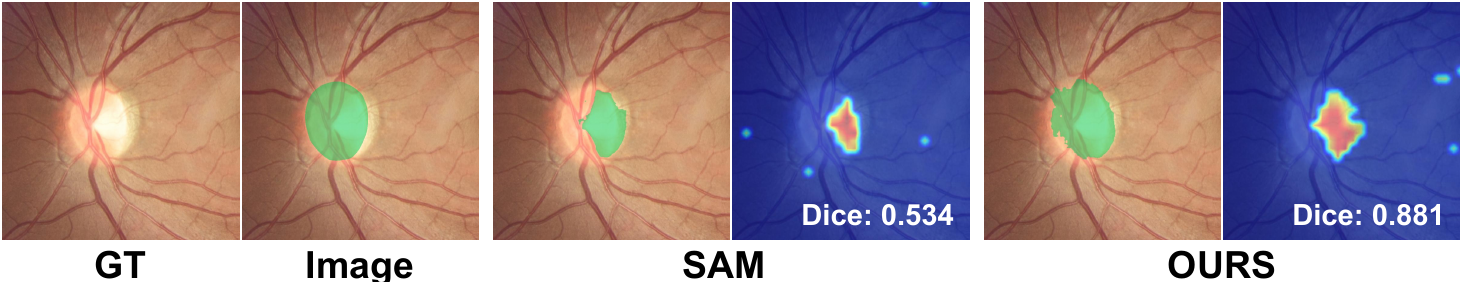} 
\caption{
Comparison of Grad-CAM visualizations between the original SAM and the proposed prompt injection strategy. Samples are taken from the REFUGE dataset. 
}
\label{fig:grad-cam}
\end{figure}

Unlike conventional prompt strategies that only interact with image features at the decoder stage, our approach introduces prompt-image interaction early in the encoder. This early fusion allows feature enhancement to directly influence multi-stage feature extraction, enabling the encoder to learn prompt-aware representations from the beginning. 
Fig.~\ref{fig:grad-cam} illustrates that the encoder becomes more sensitive to target regions, improving localization accuracy and robustness in challenging medical images. 

\subsection{Cross-Layer Boundary-Aware Attention Alignment} 
\label{subsec:3.3boundaryAlign}

To address the issue of attention dilution in the deep layers of the SAM image encoder caused by excessively large receptive fields, we design a cross-layer boundary alignment mechanism. This approach leverages the superior edge-capturing ability of shallow features to constrain the deep features, thereby enhancing boundary localization and mitigating the loss of edge information.

To obtain reliable boundary supervision, we first compute an boundary-aware map $\mathbf{M} \in \mathbb{R}^{B \times 1 \times H \times W}$ from the shallow features. Considering that the Sobel operator is computationally efficient and capable of providing accurate edge cues, we adopt it to highlight boundary regions. The construction process of $\mathbf{M}$ is detailed in Eq.~\ref{eq:sobel1} and Eq.~\ref{eq:sobel2}.

\begin{equation}
\label{eq:sobel1}
    \mathbf{G}_x = \mathbf{F}^{(l_s)} \ast \mathbf{K}_x, \quad  
    \mathbf{G}_y = \mathbf{F}^{(l_s)} \ast \mathbf{K}_y
\end{equation}

\begin{equation}
\label{eq:sobel2}
    \mathbf{M} = \frac{\sqrt{\mathbf{G}_x^2 + \mathbf{G}_y^2}}{\max(\sqrt{\mathbf{G}_x^2 + \mathbf{G}_y^2}) + \epsilon}
\end{equation}
Let $\mathbf{F}^{(l_s)}$ and $\mathbf{F}^{(l_d)}$ denote the shallow and deep feature maps, where $l_s$ and $l_d$ index the selected encoder layers. The predefined Sobel kernels $\mathbf{K}_x,\mathbf{K}_y \in \mathbb{R}^{3 \times 3}$ are used to compute the horizontal and vertical gradients in \eqref{eq:sobel1}. The boundary-aware map $\mathbf{M} \in \mathbb{R}^{B \times 1 \times H \times W}$ is then obtained by normalizing the gradient magnitude as in \eqref{eq:sobel2}, where $\epsilon>0$ ensures numerical stability.

Given the boundary-aware map $\mathbf{M} \in \mathbb{R}^{B \times 1 \times H \times W}$, the shallow and deep boundary features are extracted by masking, and the Pearson correlation between these features within the boundary regions is computed to formulate the alignment loss. The related formulations are given as follows:

\begin{equation}
\label{eq:feature}
    \mathbf{F}_s = \mathbf{F}^{(l_s)} \odot \mathbf{M}, \quad
    \mathbf{F}_d = \mathbf{F}^{(l_d)} \odot \mathbf{M}.
\end{equation}

\begin{equation}
\label{eq:pearson}
    r_{sd}^{bc} = 
    \frac{\sum\limits_{(i,j)\in \mathbf{M}} \left(\mathbf{F}_{s}^{bc}(i,j) - \mu_{s}^{bc}\right)\left(\mathbf{F}_{d}^{bc}(i,j) - \mu_{d}^{bc}\right)}
    {\sqrt{\sum\limits_{(i,j)\in \mathbf{M}} \left(\mathbf{F}_{s}^{bc}(i,j) - \mu_{s}^{bc}\right)^{2}} \, 
     \sqrt{\sum\limits_{(i,j)\in \mathbf{M}} \left(\mathbf{F}_{d}^{bc}(i,j) - \mu_{d}^{bc}\right)^{2}}}
\end{equation}

\begin{equation}
\label{eq:boundary_align}
    \mathcal{L}_{\mathrm{align}} = \frac{1}{BC} \sum_{b=1}^{B} \sum_{c=1}^{C} \left( 1 - r_{sd}^{bc} \right)
\end{equation}
We use $\mathbf{F}_s, \mathbf{F}_d \in \mathbb{R}^{B \times C \times H \times W}$ represent the masked feature maps obtained by applying the element-wise product $\odot$ with $\mathbf{M}$. The superscripts $b$ and $c$ in $r_{sd}^{bc}$ indicate the $b$-th sample and the $c$-th channel, respectively. The terms $\mu_{s}^{bc}$ and $\mu_{d}^{bc}$ denote the mean values of $\mathbf{F}_{s}^{bc}$ and $\mathbf{F}_{d}^{bc}$ within the masked boundary region.

Moreover, we only compute the alignment loss when the boundary map $\mathbf{M}$ is of high quality to ensure stability. We define a high-quality boundary map as one whose average activation within a target box significantly exceeds the average activation outside the box:

\begin{equation}
\label{eq:boundary_align_check}
    \text{mean}_{\text{in box}} > \tau \cdot \text{mean}_{\text{outside}}, \quad \text{and} \quad \text{mean}_{\text{in box}} > \delta
\end{equation}

This cross-layer boundary-aware attention alignment not only improves object boundary localization accuracy but also preserves semantically meaningful spatial information across multi-scale feature representations, thereby producing more coherent and precise segmentation results. 
Furthermore, since the alignment computation and gradient backpropagation are restricted to sparse, well-localized boundary regions via $\mathbf{M}$, the proposed approach significantly reduces computational cost to $\tfrac{|\mathbf{M}|}{H \times W}$ of standard backpropagation, where $H$ and $W$ denote the height and width of the feature map. This efficiency makes the method particularly suitable for real-time applications and deployment in resource-constrained environments. 

\section{Experiment} 
In this section, we present comprehensive experiments to evaluate the effectiveness and efficiency of the proposed BA-TTA-SAM. Section \ref{sec:sec4.1datasets-metrices} describes the benchmark datasets, evaluation metrics, and implementation details. To further demonstrate the generalization ability of our model, Section \ref{sec:sec4.2competing-methods} introduces several state-of-the-art segmentation models and test-time adaptation baselines. 
Moreover, Section \ref{sec:sec4.3comparision-with-sota} and Section \ref{sec:sec4.4visual_comparision} presents quantitative comparisons and qualitative visualization analysis across multiple medical image datasets respectively. 
We further analyze and compare the inference time among various TTA methods in Section \ref{sec:sec4.5inference-time}. 
Finally, Section \ref{sec:sec4.6ablation-study} reports ablation studies to analyze the contribution of each component in our framework. 

\begin{table*}[!htb]
\caption{Comparison between BA-TTA-SAM, fully supervised models and SAM-based models. Bold numbers indicate the best test-time adaptation results.}
\centering
\footnotesize
\setlength{\tabcolsep}{8pt}
\begin{tabular}{lcccccccccccc}
\toprule
\multirow{2}{*}{Models} 
& \multicolumn{2}{c}{ISIC2017 \cite{Gutman2016SkinLA}}
& \multicolumn{2}{c}{Kvasir \cite{Jha2019KvasirSEGAS}}
& \multicolumn{2}{c}{BUSI \cite{ALDHABYANI2020104863}}
& \multicolumn{2}{c}{REFUGE-CUP \cite{orlando2020refuge}}
& \multicolumn{2}{c}{REFUGE-DISC \cite{orlando2020refuge}}
& \multicolumn{2}{c}{Average} \\
\cmidrule(lr){2-3} \cmidrule(lr){4-5}
\cmidrule(lr){6-7} \cmidrule(lr){8-9}
\cmidrule(lr){10-11} \cmidrule(lr){12-13}
& Dice & mIOU
& Dice & mIOU
& Dice & mIOU
& Dice & mIOU 
& Dice & mIOU
& Dice & mIOU \\
\midrule
\multicolumn{13}{c}{\textbf{Fully supervised}} \\
\midrule
U-Net \cite{Ronneberger2015UNetCN}& 0.852 & 0.768 & 0.882 & 0.815 & 0.782 & 0.702 & 0.707 & 0.566 & 0.863 & 0.771 & 0.817 & 0.724 \\
U-Net++ \cite{Zhou2018UNetAN}& 0.857 & 0.776 & 0.890 & 0.826 & 0.785 & 0.710 & 0.702 & 0.564 & 0.851 & 0.749 & 0.817 & 0.725 \\
TransUNet \cite{Heidari2022HiFormerHM}& 0.857 & 0.774 & 0.892 & 0.832 & 0.776 & 0.698 & 0.693 & 0.549 & 0.869 & 0.780 & 0.817 & 0.727 \\
HiFormer-B \cite{Heidari2022HiFormerHM}& 0.865 & 0.786 & 0.861 & 0.791 & 0.752 & 0.665 & 0.698 & 0.571 & 0.821 & 0.712 & 0.799 & 0.705 \\
\midrule
\multicolumn{13}{c}{\textbf{SAM-based}} \\
\midrule
SAM (zero-shot) \cite{kirillov2023segment}& 0.758 & 0.757 & 0.797 & 0.822 & 0.816 & 0.846 & 0.752 & 0.804 & 0.743 & 0.802 & 0.773 & 0.806 \\
TENT \cite{Wang2020FullyTA}& 0.835 & 0.824 & 0.857 & 0.849 & 0.840 & 0.863 & 0.752 & 0.804 & 0.659 & 0.751 & 0.789 & 0818 \\
CoTTA \cite{Wang2022ContinualTD}& 0.819 & 0.826 & 0.827 & 0.833 & 0.818 & 0.849 & 0.745 & 0.800 & 0.700 & 0.776 & 0.782 & 0.817 \\
DELTA \cite{zhao2023delta}& 0.776 & 0.797 & 0.849 & 0.846 & 0.826 & 0.856 & 0.753 & 0.805 & 0.744 & 0.802 & 0.790 & 0.821 \\
\textbf{OURS (BA-TTA)} & \textbf{0.885} & \textbf{0.874} & \textbf{0.862} & \textbf{0.851} & \textbf{0.900} & \textbf{0.901} & \textbf{0.907} & \textbf{0.915} & \textbf{0.931} & \textbf{0.935} & \textbf{0.897} & \textbf{0.895} \\
\midrule
WeSAM \cite{10657214}& 0.872 & 0.777 & 0.832 & 0.757 & 0.899 & 0.819 & 0.868 & 0.769 & 0.912 & 0.839 & 0.877 & 0.792 \\
MSA \cite{WU2025103547}& 0.923 & 0.862 & 0.914 & 0.854 & 0.908 & 0.836 & 0.782 & 0.653 & 0.822 & 0.705 & 0.870 & 0.782 \\
MedSAM \cite{Ma2023SegmentAI}& 0.924 & 0.863 & 0.902 & 0.831 & 0.916 & 0.848 & 0.870 & 0.773 & 0.940 & 0.888 & 0.910 & 0.841 \\
\bottomrule
\end{tabular}
\vspace{-3mm}
\label{tb:competing_sota}
\end{table*}

\subsection{Experimental Settings}
\label{sec:sec4.1datasets-metrices}
\subsubsection{Datasets}
To comprehensively evaluate the effectiveness and generalization ability of the proposed method across diverse medical segmentation tasks, we conduct experiments on four publicly available datasets. These datasets span dermoscopic, ultrasound, endoscopic, and fundus image domains.

\textbf{ISIC2017 \cite{Gutman2016SkinLA}:}
This is a dermoscopic skin lesion segmentation dataset consisting of 2,000 training, 150 validation, and 600 test images, following the official split. It is widely used for melanoma segmentation benchmarking.

\textbf{Kvasir-SEG \cite{Jha2019KvasirSEGAS}:}
This is a gastrointestinal polyp segmentation dataset with 1,000 high-resolution endoscopic images. The dataset is randomly split into 770 training, 110 validation, and 120 testing samples.

\textbf{BUSI \cite{ALDHABYANI2020104863}:}
This dataset contains 780 breast ultrasound images, including 437 benign and 210 malignant tumors, and 133 normal cases (excluded due to lack of diagnostic information). We use the remaining 647 tumor images and perform a stratified 6:2:2 split, resulting in 388 training, 129 validation, and 130 test images.

\textbf{REFUGE \cite{orlando2020refuge}:}
The REFUGE dataset contains 1,200 retinal fundus images with annotations for optic disc and cup segmentation. In this work, we follow the official data split and use 400 images each for training, validation, and testing.

Note that the above-mentioned training and validation splits are utilized only in experiments involving fully supervised and SAM-based models with fine-tuning, not in test-time adaptation setting. 

\subsubsection{Evaluation Metrics}
To comprehensively evaluate the segmentation performance, we adopt the Dice Similarity Coefficient (DSC) and the mean Intersection over Union (mIoU) as our evaluation metrics. These two metrics jointly assess regional overlap, boundary accuracy, and prediction consistency across the spatial domain.
The DSC measures the spatial overlap between the predicted segmentation mask and the ground-truth mask and the mean Intersection over Union (mIoU) quantifies the pixel-wise overlap between prediction and ground-truth. 

\subsubsection{Implementation Details}
\label{sec:sec4.1.2implementation-details}
All experiments are implemented using the PyTorch framework. Models that require fine-tuning are trained for 100 epochs and the initial learning rate is set to 1e-4, with the best-performing checkpoint on the validation set selected for final evaluation on the test set. 
For prompt generation during training and inference, we adopt a strategy to simulate practical segmentation scenarios. Specifically, we generate a bounding box prompt that minimally covers the entire object area. This approach mimics user interaction while ensuring robust generalization across varying medical image modalities. 

\subsection{Competing Methods}
\label{sec:sec4.2competing-methods}
We evaluate our approach against three types of baselines: fully supervised segmentation models, fine-tuned SAM and test-time adaptation methods in Table~\ref{tb:competing_sota}. This comprehensive comparison framework enables an extensive evaluation across multiple supervision levels and domain adaptation strategies.

\subsubsection{Fully-supervised}
We include four widely-used fully supervised segmentation models as strong baselines: 
\textbf{U-Net \cite{Ronneberger2015UNetCN}} is a classic encoder–decoder architecture with skip connections, widely adopted in medical image segmentation tasks.
\textbf{U-Net++ \cite{Zhou2018UNetAN}} enhances U-Net with nested and dense skip pathways to improve multi-scale feature aggregation.
\textbf{TransUNet \cite{Chen2021TransUNetTM}} integrates a Transformer encoder with a U-Net decoder to capture long-range dependencies while preserving detailed spatial information for medical image segmentation. 
\textbf{HiFormer \cite{Heidari2022HiFormerHM}} is a hybrid architecture that integrates CNNs and Swin Transformers with a double-level fusion module to capture both local and global features for medical image segmentation. 

\subsubsection{SAM-based}
We evaluate the original SAM without any adaptation to establish a zero-shot baseline. We also include three representative test-time adaptation methods and three fine-tuned SAM-based models for comparision. 
\textbf{TENT} \cite{Wang2020FullyTA} performs test-time adaptation by optimizing the model to produce confident predictions on unlabeled target data through entropy minimization during inference.
\textbf{CoTTA} \cite{Wang2022ContinualTD} combines mean teacher and temporal dropout to stabilize predictions on new domains. 
\textbf{DELTA} \cite{zhao2023delta} employs renormalization and dynamic re-weighting to correct unreliable feature statistics and mitigate class bias. 
\textbf{WeSAM} \cite{10657214} updates low-rank weights for adapting SAM to medical imaging tasks. 
\textbf{MSA} \cite{WU2025103547} introduces lightweight domain-aware adapters into SAM to align its representations with medical image characteristics. 
\textbf{MedSAM} \cite{Ma2023SegmentAI} fine-tunes SAM end-to-end on medical image-mask datasets as a full fine-tuning variant. 
All methods compared employ the ViT-B backbone inherited from SAM. 

\begin{figure*}[!htb]
\centering
\includegraphics[width=1.0\textwidth]{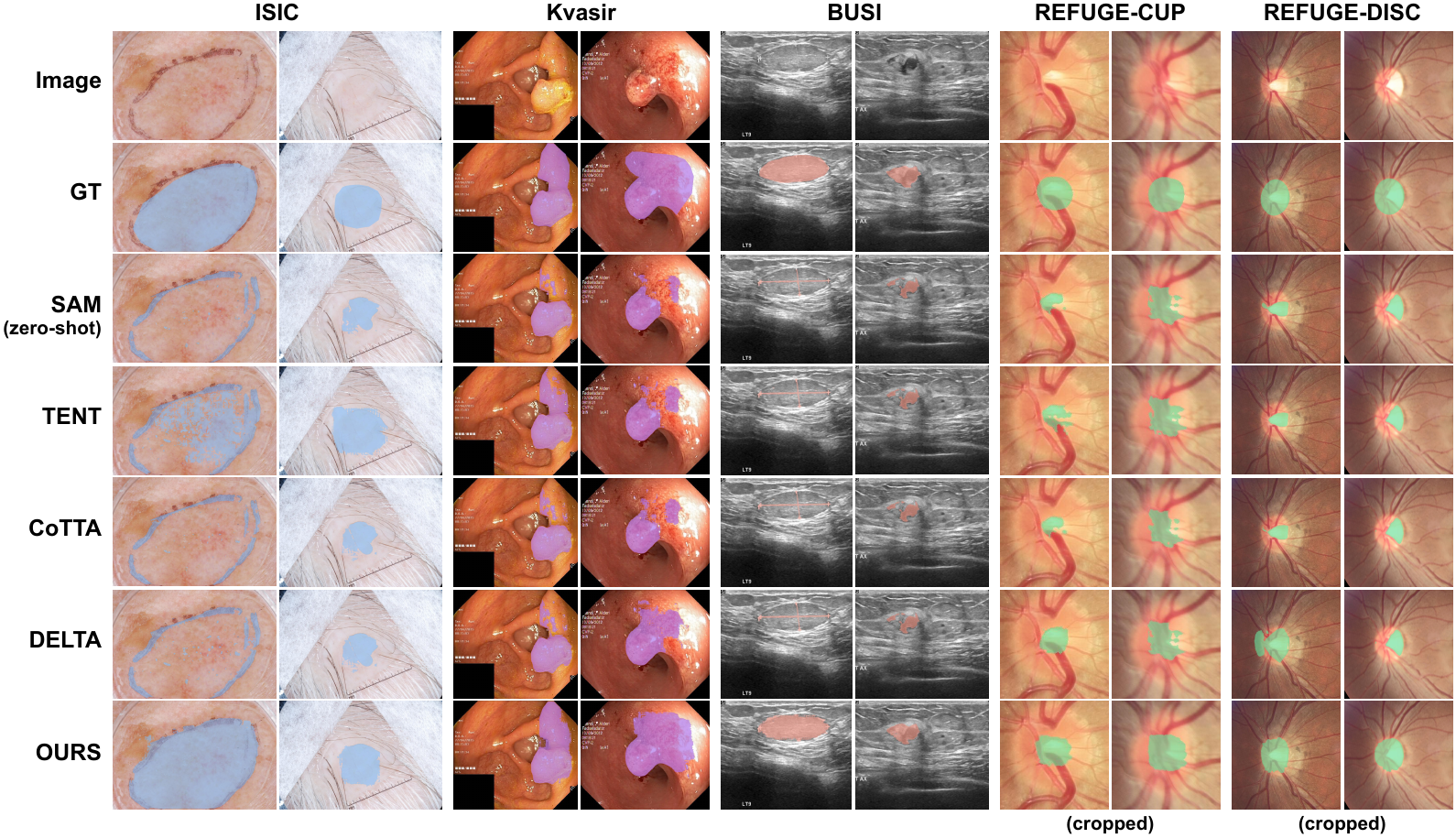} %
\caption{
Qualitative visualization of segmentation results across challenging medical datasets. The comparison includes SAM, existing TTA baselines, and our proposed BA-TTA-SAM. 
}
\label{fig:visual_comparision}
\end{figure*}

\subsection{Comparison with State-of-the-art}
\label{sec:sec4.3comparision-with-sota}
To further evaluate the generalizability of the proposed strategy, we conduct experiments across a diverse set of representative baselines. Specifically, we select four widely used fully-supervised segmentation models and several SAM-based variants. This setup allows us to assess the segmentation performance of mainstream paradigms under consistent evaluation protocols in the medical imaging domain. 

As shown in Table \ref{tb:competing_sota}, fully supervised models generally outperform SAM in zero-shot settings but still fall short of the performance achieved by fine-tuned SAM-based models, demonstrating the benefits of task-specific adaptation and  the strong segmentation capability of SAM. 
Existing test-time adaptation methods, such as TENT \cite{Wang2020FullyTA}, CoTTA \cite{Wang2022ContinualTD}, and DELTA \cite{zhao2023delta}, provide only limited improvements. On challenging datasets such as REFUGE, where the lesion regions are particularly small, these methods struggle to improve the segmentation performance of the original SAM and may even produce misleading predictions. 
The original SAM achieves an average Dice score of 77.3\% and mIoU of 80.6\% on four unseen medical image datasets. In contrast, our proposed method achieves the best performance among all TTA methods, reaching 89.7\% Dice and 89.5\% mIoU, even outperforming parameter-efficient fine-tuning methods such as WeSAM and MSA, and approaching the upper bound set by fully fine-tuned MedSAM (91.0\% Dice), while surpassing it in mIoU (84.1\%). This demonstrates that our method exhibits consistently superior performance across multiple datasets, significantly outperforming existing test-time adaptation approaches.

\begin{table*}[!htb]
\caption{
Ablation study results on the ISIC Dataset. GP-injection denotes the encoder-level Gaussian prompt injection into the image encoder. BA-Alignment signifies the boundary-aware attention alignment in the image encoder.
}
\centering
\setlength{\tabcolsep}{4pt} 
\small
\begin{tabular}{ccccccccccccccc}
\toprule
\multirow{2}{*}{SAM} & \multirow{2}{*}{GP-Injection} & \multirow{2}{*}{BA-Alignment} & \multicolumn{2}{c}{ISIC} & \multicolumn{2}{c}{Kvasir} & \multicolumn{2}{c}{BUSI} & \multicolumn{2}{c}{REFUGE\_CUP} & \multicolumn{2}{c}{REFUGE\_DISC} & \multicolumn{2}{c}{Average}\\
\cmidrule{4-15}
    & & & Dice & mIOU & Dice & mIOU & Dice & mIOU & Dice & mIOU & Dice & mIOU & Dice & mIOU\\
\midrule
$\checkmark$ & & & 0.758 & 0.757 & 0.797 & 0.822 & 0.816 & 0.846 & 0.752 & 0.804 & 0.743 & 0.802 & 0.733 & 0.806 \\
$\checkmark$ & $\checkmark$ & & 0.877 & 0.868 & 0.862 & 0.851 & 0.896 & 0.899 & 0.864 & 0.881 & 0.903 & 0.913 & 0.880 & 0.882 \\
$\checkmark$ & & $\checkmark$ & 0.776 & 0.797 & 0.853 & 0.847 & 0.826 & 0.856 & 0.752 & 0.804 & 0.746 & 0.803 & 0.791 & 0.821 \\
$\checkmark$ & $\checkmark$ & $\checkmark$ & 0.885 & 0.874 & 0.862 & 0.851 & 0.900 & 0.901 & 0.907 & 0.915 & 0.931 & 0.935 & \textbf{0.897} & \textbf{0.895} \\
\bottomrule
\end{tabular}
\label{tb:ablation_study}
\end{table*}

\subsection{Qualitative Visualization}
\label{sec:sec4.4visual_comparision}
Fig.~\ref{fig:visual_comparision} presents the qualitative segmentation results of different models on four challenging medical image datasets. As illustrated, the original SAM is sensitive to noise commonly found in medical images under zero-shot settings. 
SAM and existing TTA methods tend to overemphasize boundary and intensity cues, often segmenting only salient edges and manual annotation noise, as evidenced by the first examples from ISIC and BUSI. 
Furthermore, additional examples indicate that these methods frequently produce partial lesion segmentation. Representative cases from REFUGE-DISC demonstrate that the models segment only the bright subregions while missing the full lesion extent. 
Overall, existing test-time adaptation (TTA) methods do not effectively mitigate the domain shift faced by SAM in zero-shot scenarios, achieving performance comparable to the original SAM. 

In contrast, our proposed BA-TTA-SAM effectively addresses these issues. As shown in the figure, our method consistently identifies the target regions with higher accuracy across datasets, yielding clearer and more precise segmentation results. 
Benefiting from the proposed Gaussian Prompt Injection and Boundary-Aware Attention Alignment mechanisms, our model receives enhanced spatial guidance and boundary awareness during the encoding stage. This leads to significantly improved performance in complex medical imaging conditions and greater robustness against challenging or degraded cases. 

\subsection{Inference Time}
\label{sec:sec4.5inference-time}
\begin{figure}[htb]
\centering
\includegraphics[width=.8\columnwidth]{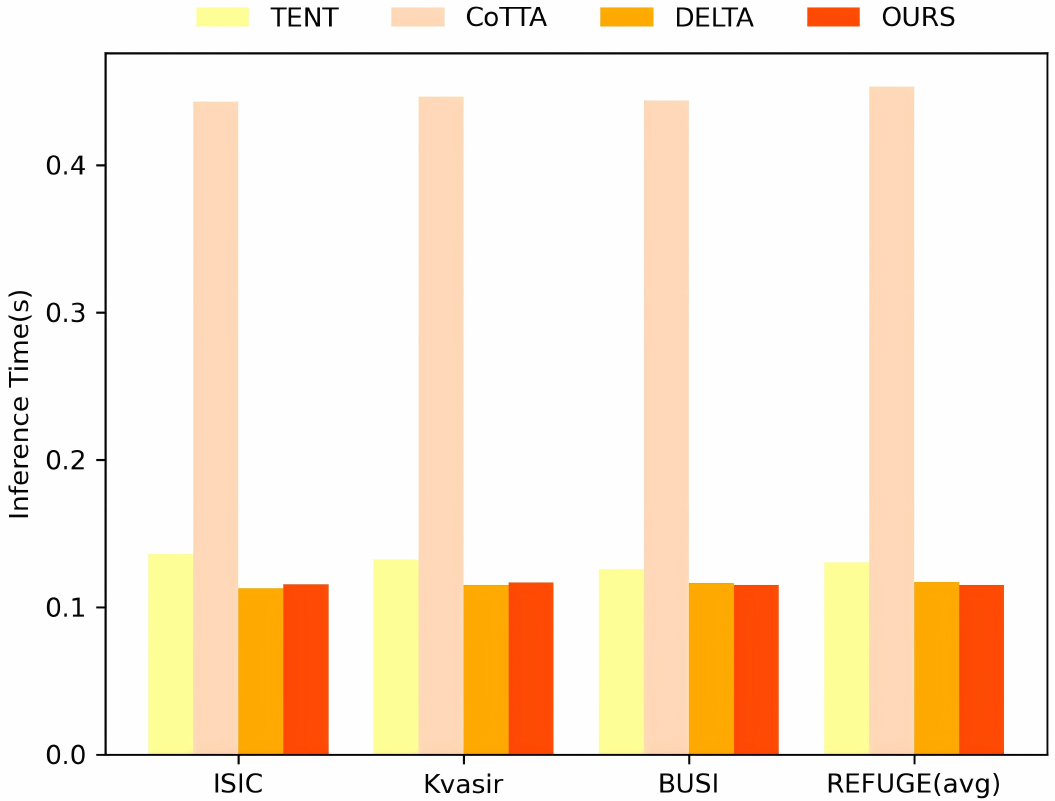} %
\caption{
Comparison of inference time. 
Bar heights indicate the average inference time per image, computed over the ISIC, Kvasir, BUSI, and REFUGE datasets. 
}
\label{fig:inference_time_conparison}
\end{figure}

Fig.~\ref{fig:inference_time_conparison} summarizes the average inference time per image for different TTA methods. CoTTA exhibits the longest inference time, as it requires multiple forward passes per input to generate high-quality pseudo-labels via mean teacher and augmented predictions, and randomly restores a subset of neurons during inference. DELTA and our method have comparable times, slightly lower than TENT. Notably, our method achieves the best performance while maintaining low inference time, demonstrating the efficiency and practical advantage of the BA-TTA-SAM framework. 

\subsection{Ablation Study}
\label{sec:sec4.6ablation-study}
To verify the synergistic benefits of the Gaussian prompt injection and the boundary-aware attention alignment in medical image segmentation, this section presents integrated experiments evaluating their combined performance in different datasets. 

As shown in TABLE \ref{tb:ablation_study}, incorporating Gaussian heatmap-based prompt injection leads to a notable improvement in both DICE and mIoU metrics. 
Specifically, the proposed prompt injection strategy yields significant DICE improvements across four datasets, with gains of 11.9\%, 6.5\%, 8.0\%, 11.2\%, and 16.0\%, respectively.
In addition, our boundary-aware attention alignment, which only optimizes a small number of parameters, also demonstrates strong performance, increasing the average DICE from 73.3\% to 79.1\% across the four datasets.
When combined, the two strategies lead to an overall improvement of 16.4\% in average DICE and 8.9\% in average mIoU across all datasets.

Results show that both individual strategies improve performance, while the combined configuration achieves the best segmentation accuracy across all models and datasets. 
The integrated strategy combines the spatial guidance of boundary-aware attention alignment with the semantic adaptability of Gaussian heatmap-based prompt injection, improving parameter efficiency while enhancing segmentation accuracy during inference.

\section{Discussion} 
\label{sec:discussion}
Recent test-time adaptation (TTA) studies \cite{Wang2020FullyTA, zhao2023delta, iwasawa2021testtime} have primarily focused on addressing the distribution shift between the source domain and the target domain, while often overlooking the intrinsic architectural characteristics of the model itself. 
For instance, the Segment Anything Model (SAM) inherently contains prompt-based prior information, which has not been fully exploited, and the stacked Vision Transformer layers exhibit complementary functional roles. Shallow and deep layers attend to different aspects of visual features, providing complementary cues. 
In this work, we proposed a test-time adaptation framework based on SAM. Comprehensive experiments are conducted on four medical image datasets from different medical domains, and the results consistently demonstrate the superiority of our method. The proposed model mainly consists of two components. 
In particular, Section \ref{secsec:prompt_injection_strategy} and Section \ref{secsec:prompt_injection_ablation_study} present a more detailed discussion and analysis of the prompt injection strategy and the influence of injection depth on model performance. 
Finally, we discuss the limitation of our current work and potential future directions in Section \ref{secsec:limitation_and_future_work}.  
\subsection{Prompt Injection Strategy}
\label{secsec:prompt_injection_strategy}
To effectively integrate prompt information into the image encoder and enhance the model's focus on key regions, this paper systematically designs and compares three types of prompt injection strategies, using SAM as the backbone. The last two types are further divided into 'Feature Enhancement' and 'Spatial Prior' variants, corresponding to the prompt injection locations and the pentagram-marked positions in Fig.~\ref{fig:Frame2}, respectively.

\subsubsection{Image Overlay} 
This strategy directly overlays prompt information (e.g., points, boxes) onto the input image through pixel modifications to guide the model to perceive the prompt location. 
However, this method is easily affected by factors such as image resolution and color distribution, limiting its robustness.

\subsubsection{Embedding-based Injection} 
In this strategy, the prompt is encoded into embeddings and used in two ways. First, it is injected before each Transformer block as feature enhancement. Second, it is inserted after the attention layer to refine the spatial distribution of attention weights. 

\subsubsection{Gaussian-based Injection} 
The prompt is transformed into Gaussian heatmaps and utilized in two ways, consistent with the embedding-based strategy: as feature enhancement added before each Transformer block and as a spatial prior injected after the attention layer to steer attention distribution. 
%

Experimental results show that the feature enhancement in the Gaussian-based injection strategy performs best. 
By enriching feature diversity, it significantly improves segmentation accuracy. 
This strategy offers superior focus on target regions and effectively reduces background interference. A summary of the results is presented as Table \ref{tb:prompt-injection-comparision}.

\begin{table}[!htb]
\caption{Dice score comparison of different prompt injection strategies. 
A: Prompts are injected as feature enhancement. B: Prompts are injected as spatial prior. 
}
\centering
\setlength{\tabcolsep}{1pt} 
\small
\begin{tabular}{llccccc}
\toprule
    & Dataset & Overlay & \multicolumn{2}{c}{Embedding-based} & \multicolumn{2}{c}{Gaussian-based}\\
    \cmidrule(lr){4-5} \cmidrule(lr){6-7} %
    & &  & A & B & A & B \\
\midrule
\multirow{4}{*}{\centering Point}
    & \centering ISIC & 0.692 & 0.683 & 0.688 & \textbf{0.726} & 0.664 \\
    & \centering Kvasir & 0.811 & 0.777 & 0.800 & \textbf{0.834} & 0.728 \\
    & \centering BUSI & 0.676 & 0.713 & 0.732 & \textbf{0.765} & 0.686 \\
    & \centering REFUGE (avg)& 0.535 & 0.509 & 0.527 & \textbf{0.591} & 0.556 \\
\midrule
\multirow{4}{*}{\centering Box}
    & \centering ISIC & 0.809 & 0.761 & 0.759 & \textbf{0.877} & 0.874 \\
    & \centering Kvasir & 0.750 & 0.796 & 0.796 & \textbf{0.862} & 0.781 \\
    & \centering BUSI & 0.776 & 0.816 & 0.811 & \textbf{0.896} & 0.868 \\
    & \centering REFUGE (avg)& 0.873 & 0.752 & 0.749 & 0.884 & \textbf{0.897} \\
\bottomrule
\end{tabular}
\label{tb:prompt-injection-comparision}
\end{table}
    
\subsection{Prompt Injection Ablation Study}
\label{secsec:prompt_injection_ablation_study}
To investigate the impact of injection depth in SAM, we conducted ablation experiments comparing different stage configurations. Our setup adopts ViT-B, where the image\_encoder contains 12 blocks grouped into stages of 3 blocks each. Within each stage, the first two blocks employ window-based attention mechanisms, while the last one uses self-attention. Results indicate that more injections generally yield more significant performance improvements. However, on certain small-scale datasets, the stage*3 injection strategy demonstrates better balance, as shown in Table \ref{tb:prompt_injection}.. It validates that the number of stages should be adapted to practical conditions, which points to a potential direction for future work. 

\begin{table}[!htb]
\caption{Dice score comparison of the encoder-level Gaussian prompt injection with different numbers of stages.}
\centering
\setlength{\tabcolsep}{1pt} 
\begin{tabular}{llcccccc@{}}
\toprule
 & Dataset & Base & Stage*1 & Stage*2 & Stage*3 & Stage*4 \\
\midrule
\multirow{4}{*}{Point} 
    & ISIC & 0.683 & 0.711 & 0.722 & \textbf{0.727} & 0.726 \\
    & Kvasir & 0.791 & 0.822 & 0.826 & \textbf{0.836} & 0.834 \\
    & BUSI & 0.735 & 0.737 & 0.752 & 0.761 & \textbf{0.765} \\
    & REFUGE (avg)& 0.532 & 0.538 & 0.554 & 0.581 & \textbf{0.591} \\
\midrule  
\multirow{4}{*}{Box} 
    & ISIC & 0.758 & 0.803 & 0.841 & 0.864 & \textbf{0.877} \\
    & Kvasir & 0.797 & 0.808 & 0.817 & 0.833 & \textbf{0.862} \\
    & BUSI & 0.816 & 0.839 & 0.869 & 0.886 & \textbf{0.896} \\
    & REFUGE (avg)& 0.748 & 0.791 & 0.829 & 0.859 & \textbf{0.884} \\
\bottomrule
\end{tabular}
\label{tb:prompt_injection}
\end{table}

\subsection{Limitations and Future work}
\label{secsec:limitation_and_future_work}
As shown in Table \ref{tb:prompt_injection}, when the input is the point prompt, injecting Gaussian prompts into only three stages sometimes achieves comparable or even better results than injecting into all four stages. This observation suggests that different layers exhibit varying sensitivities to Gaussian prompt injection. 
In the future, exploring the cross-layer sensitivity relationships and functional differentiation among layers is expected to further improve model performance. 

\section{Conclusion}
In this paper, we present BA-TTA-SAM, a novel task-agnostic test-time adaptation framework specifically designed to leverage the architectural strengths of Vision Transformers. The proposed method effectively mitigates performance degradation caused by distribution shifts between pretraining datasets (e.g., SA-1B) and medical imaging domains. Extensive experiments across multiple medical image segmentation benchmarks demonstrate that our framework significantly surpasses existing test-time adaptation approaches. 
Notably, it achieves competitive or even superior performance compared to models with high computational cost, such as WeSAM, MSA, and MedSAM. 
This framework effectively enhances SAM's adaptability to downstream medical tasks, paving the way for robust, cost-effective deployment of foundation models in clinical applications. 

\section*{Declaration of interest statement}
The authors declare that they have no known competing financial interests or personal relationships that could have appeared to influence the work reported in this paper. 

\section*{Ethics statement}
This study used only publicly available medical image datasets. 
Data collection, analysis, and reporting followed strict principles of accuracy and integrity, without any data fabrication or falsification. The research is original, and all authors contributed substantially and ethically, with no plagiarism or duplicate submission involved. 

\section*{Acknowledgements}
This work was supported by the National Natural Science Foundation of China under Grant 62376174, Grant 62025601, and Grant U24A20341. 




\bibliographystyle{elsarticle-num}  
\bibliography{ref}                  

\end{document}